%% file: main.tex
\pdfoutput=1 
\documentclass[10pt,twocolumn,letterpaper]{article}

\usepackage{iccv}
\usepackage{times}
\usepackage{graphicx}
\usepackage{amsmath}
\usepackage{amssymb}
\usepackage{graphicx}
\usepackage{booktabs}
\usepackage{multirow}
\usepackage{subcaption}
\usepackage[algo2e,ruled,vlined]{algorithm2e}
\usepackage{algorithm}
\usepackage{stfloats}

\newcommand{\cutsubsectionup}{\vspace*{-0.00in}}
\newcommand{\cutparagraphup}{\vspace*{-0.1in}}

\usepackage[pagebackref=true,breaklinks=true,letterpaper=true,colorlinks,bookmarks=false]{hyperref}

\iccvfinalcopy 

\def\iccvPaperID{9186} 

\ificcvfinal\fi
\DeclareMathOperator*{\argmin}{arg\,min}
\DeclareMathOperator*{\argsort}{arg\,sort}

\begin{document}

\title{Towards Attack-tolerant Federated Learning via Critical Parameter Analysis}

\input{authors}

\maketitle
\ificcvfinal\thispagestyle{empty}\fi

\newcommand{\model}{\textsf{FedCPA}}
\input{contents/0_abstract}
\input{contents/1_introduction}
\input{contents/2_related_works}
\input{contents/3_problem_statement}
\input{contents/4_discovering_pattern}
\input{contents/5_model}

\input{contents/6_result}

\input{contents/7_conclusion}

{\small
\bibliographystyle{ieee_fullname}
\bibliography{egbib}
}
\newpage
\input{contents/8_appendix}

\end{document}

%% file: authors.tex
\author{Sungwon Han\thanks{Equal contribution to this work.}\textsuperscript{\rm ~~1}~~~Sungwon Park\footnotemark[1]\textsuperscript{\rm ~~1}~~~Fangzhao Wu\textsuperscript{\rm 2}~~~Sundong Kim\textsuperscript{\rm 3} \\ Bin Zhu\textsuperscript{\rm 2}~~~Xing Xie\textsuperscript{\rm 2}~~~Meeyoung Cha\textsuperscript{\rm 4,1} \\
\textsuperscript{\rm 1} KAIST 
\hspace{0.3cm} \textsuperscript{\rm 2} Microsoft Research Asia \hspace{0.3cm} 
\textsuperscript{\rm 3} GIST
\hspace{0.3cm} \textsuperscript{\rm 4} Institute for Basic Science 
}

%% file: contents/0_abstract.tex
\begin{abstract}
Federated learning is used to train a shared model in a decentralized way without clients sharing private data with each other.
Federated learning systems are susceptible to poisoning attacks when malicious clients send false updates to the central server.
Existing defense strategies are ineffective under non-IID data settings.
This paper proposes a new defense strategy, \model{} \textsf{(Federated learning with Critical Parameter Analysis)}. Our attack-tolerant aggregation method is based on the observation that benign local models have similar sets of top-$k$ and bottom-$k$ critical parameters, whereas poisoned local models do not.
Experiments with different attack scenarios on multiple datasets demonstrate that our model outperforms existing defense strategies in defending against poisoning attacks.
\looseness=-1
\end{abstract}

%% file: contents/1_introduction.tex
\section{Introduction}

\noindent
The proliferation of computing devices like mobile phones has led to an increase in proprietary user data.
The abundance of user data offers the opportunity to create numerous applications but also raises concerns about data privacy.
Federated learning (FL) is a cutting-edge collaborative technique that addresses the privacy challenge by enabling machine learning on decentralized devices without exchanging locally stored data~\cite{wang2021field}.
For example, a prominent FL model, FedAvg~\cite{mcmahan2017communication}, works as follows: Given a central server and multiple clients, the central server selects a random subset of clients and sends the global model to them.
Then, each selected client uses its own data to optimize the local model and sends back the model update to the central server.
The central server takes the \textit{average} of these received updates to construct a new global model.
This FL framework enables a decentralized system to train a globally shared model via aggregating updates from local models while preserving data privacy.

However, the averaging operation used in the central server leaves room for \textit{poisoning attacks}~\cite{baruch2019little,lyu2020threats} when malicious clients pose as ordinary clients and submit fraudulent model updates.
Attackers can not only impede the convergence of model training and degrade performance~\cite{steinhardt2017certified} (which is called \emph{untargeted attacks}) but they can also manipulate model updates by injecting a backdoor into the resulting global model without substantially degrading its performance~\cite{tran2018spectral} (which is called \emph{targeted attacks}). \looseness=-1

Several defense strategies have been proposed to eliminate false updates from potentially malicious clients and maintain benign updates on FL systems.
For instance, one idea is to use outlier-resistant statistics such as the median or trimmed mean~\cite{xie2018generalized,yin2018byzantine} rather than the average in model aggregation.
Blanchard et al.~\cite{blanchard2017machine} proposed Krum, which removes atypical model updates with low local density compared to their $k$-nearest neighbors.
Fung et al.~\cite{fung2020limitations} and Fu et al.~\cite{fu2019attack} proposed weighted averaging of local updates in proportion to each update's normality level.
Nevertheless, these defense strategies cannot detect adversaries in so-called non-IID (non-independent, identically distributed) situations, where data distributions vary substantially among clients.
Existing defense strategies project model updates as individual Euclidean vectors and evaluate their abnormality based on their distances from other model updates.
Meanwhile, the non-IID property leads to diverse benign updates, which makes malicious and benign updates indistinguishable in Euclidean space. As a result, existing defense strategies become ineffective~\cite{baruch2019little,park2023feddefender}. \looseness=-1

This paper presents \model{} (Federated learning with Critical Parameter Analysis), an attack-tolerant aggregation method for FL under non-IID data settings. Inspired by a recent observation that not all model parameters contribute equally to optimization~\cite{frankle2019lottery,xia2021robust}, we assess the importance of the model parameters in every client's update.
Our analysis shows that benign model updates share similar sets of top-$k$ and bottom-$k$ important parameters, even under non-IID data.
However, this pattern is not observed for malicious model updates. 
Based on this observation, we propose a new defense strategy tailored for FL systems to measure model similarity, which extends beyond the extant Euclidean-based similarity and provides an efficient way to discard updates from clients that are likely malicious. \looseness=-1

\model{} consists of two steps: (1) computing the \textit{normality} score of each client's model concerning parameter importance and (2) aggregating local updates via a weighted average to remove the effect of likely-malicious updates.
In the first step, the importance of each parameter is computed by multiplying its value by its change after local training.
The resulting parameters are then ranked in order of importance.
Top-$k$ and bottom-$k$ most important parameters are extracted for each client's model and used to compute the similarity among clients' models.
Then, we define the normality of the model update to measure its similarity with other model updates. Model updates that differ from other updates are considered malicious.
In the second step, outlier local updates are filtered out by adjusting their weights regarding their normality scores. \looseness=-1

Our evaluation demonstrates that \model{} protects against both untargeted and targeted attacks better than existing methods such as Multi-Krum~\cite{blanchard2017machine}, FoolsGold~\cite{fung2020limitations}, and ResidualBase~\cite{fu2019attack}. We make the following contributions: \looseness=-1
\begin{itemize}
    \item We empirically show that benign local models in federated learning exhibit similar patterns in how parameter importance changes during training. The top and bottom parameters have smaller rank order disruptions than the medium-ranked parameters.

    \item Based on the data observation that holds over non-IID cases, we present a new metric for measuring model similarity (Eq.~\ref{eq:model_similarity}). With this measure, \model{} can efficiently assess the normality of each local update, enabling attack-tolerant aggregation.

    \item Extensive experiments demonstrate the superiority of \model{} in terms of defense performance. For example, \model{} reduces the success rate of targeted attacks by a factor of 3 (from 51.4\% to 21.9\%) on CIFAR-10 and by a factor of 2 (from 74.6\% to 43.2\%) on TinyImageNet. \looseness=-1 
\end{itemize}
The proposed model can be used in various federated learning contexts as a more robust and attack-tolerant decentralized computing framework. Codes are available at {\url{https://github.com/Sungwon-Han/FEDCPA}}.
 

%% file: contents/2_related_works.tex
\section{Related Work}
\subsection{Model Poisoning Attacks}

\noindent
Due to its decentralized nature, federated learning is susceptible to model poisoning attacks and allows malicious clients to send harmful updates to the central server without supervision~\cite{wu2022fedattack}.
As local training data is not shared, malicious participants launch attacks without a full understanding of the entire dataset~\cite{fang2020local}.
Model poisoning attacks can be categorized into untargeted and targeted attacks. \looseness=-1

In an \textit{untargeted attack} scenario, attackers aim to indiscriminately degrade the model's overall performance across all classes~\cite{shafahi2018poison}.
Simple and widely used methods of untargeted attack include label-flipping and adding Gaussian noise, which can be executed without prior knowledge of the entire training data distribution~\cite{steinhardt2017certified}.
A label-flipping attack involves malicious clients sending false update signals by randomly altering the class label of the training data~\cite{xiao2012adversarial}.
On the other hand, Gaussian noise attacks send random noise with the same distribution as the local model prior to the attack in place of the benign client updates~\cite{fang2020local}.

In a \textit{targeted attack}, the objective of a malicious client is to deliberately introduce a backdoor into the global model, which predicts a specific target label for any input overlaid with the backdoor trigger but otherwise behaves like a normal model with a similar overall performance~\cite{bagdasaryan2020backdoor, gu2017badnets,xie2020dba}.
The backdoor trigger can be a small square to be blended into the original image or a fixed watermark on the image~\cite{chen2017targeted,liu2020reflection}.

\subsection{Defense Strategies in Federated Learning}


\noindent\textbf{Operation based strategy.}
The main objective of defense strategies is to screen harmful updates from malicious clients.
The first representative line of work involves dimension-wise aggregation, which employs outlier-resilient operations instead of a simple average.
For example, \textit{Median} aggregates local updates by computing the median value for each dimension of the updates~\cite{xie2018generalized}.
\textit{Trimmed Mean} is another aggregation method that eliminates a specified percentage of the smallest and largest values, then computes the average of the remaining values~\cite{yin2018byzantine}. \looseness=-1

When the training data is of a non-IID distribution, the median aggregation method becomes less effective because it overlooks underrepresented updates.
To tackle this limitation, \textit{RFA} suggests using an approximate geometric median operation~\cite{pillutla2022robust}.
\textit{ResidualBase}, on the other hand, introduces residual-based aggregation to determine parameter confidence after calculating the residual of each model parameter via a median estimator~\cite{fu2019attack}.
\smallskip

\noindent\textbf{Anomaly detection based strategy.}
The next line of work involves using anomaly detection to identify and remove malicious updates during aggregation.
One representative work is \textit{Krum}, which uses the Euclidean norm space to identify updates far from benign as malicious~\cite{blanchard2017machine}.
In Krum, a local model update that shows the highest similarity to $n-m-2$ of its neighboring updates is identified as benign, with $m$ denoting the anticipated number of malicious clients.
Multi-Krum extends this idea by selecting multiple benign local updates iteratively using Krum.
Another approach, \textit{FoolsGold}, identifies the coordinated actions of targeted attacks~\cite{fung2020limitations}.
Operating under the assumption that malicious clients engaged in a targeted attack exhibit similar update patterns, Foolsgold adjusts the learning rate of model updates, scaling it in proportion to the diversity of the updates.
\textit{Norm Bound} excludes clients whose local updates exceed a certain threshold for the norm, as malicious clients tend to produce updates with larger norms~\cite{sun2019can}.

%% file: contents/3_problem_statement.tex
\section{Problem Statement}
\paragraph{Federated Learning.} 
Suppose a set of $N$ clients in total in a federated learning system as $\mathcal{C}$ and a set of training sample data in the $i$-th client as $\mathcal{D}_i$ ($i\in \{1, ..., N\}$).
FL aims to train a single global model parameterized as $\phi$ without directly sharing the local dataset $\mathcal{D}_i$ with others.
Given loss objective $\mathcal{L}_{i}$ in the $i$-th client and its empirical loss $l_i$, the main objective for optimizing $\phi$ can be expressed as \looseness=-1
\begin{align}
    \argmin_\phi \mathcal{L}(\phi) &= \mathbb{E}_{i\in{[1..N]}} [\mathcal{L}_i (\phi, \mathcal{D}_i)], \nonumber \\
    \text{where } \mathcal{L}_i(\phi, \mathcal{D}_i) &= \mathbb{E}_{(\mathbf{x}, y) \in \mathcal{D}_i}[l_i (\mathbf{x}, y; \phi)]. \label{eq:fed_objective}
\end{align}

Following the literature~\cite{fung2020limitations}, we choose FedAvg~\cite{mcmahan2017communication} as the default setting to optimize Eq.~\ref{eq:fed_objective} in the following way.
FedAvg divides each training iteration into multiple steps. 
At the beginning of the $t$-th iteration ($t\geq 0$), the central server randomly selects a subset of clients and distributes its global model $\phi^{t}$.
Then, selected clients update their
local model weights $\theta_i^t$ with their dataset $\mathcal{D}_i$ and send these updates as $\Delta_i^t = \theta_i^{t} - \phi^{t}$ to the central server. 
Then, the central server aggregates receive local model updates and modifies the global model weight $\phi^{t+1}$ as follows (hereafter called the \emph{central aggregation process}):
\begin{align}
    \phi^{t+1} = \phi^t + {{\sum_{i\in [1..N]} |\mathcal{D}_i| \cdot \Delta_i^t} \over \sum_{i\in [1..N]} |\mathcal{D}_i|}. \label{eq:fedavg}
\end{align}
This process repeats until the global model converges.

\cutparagraphup
\paragraph{Threat model.}
 
Consider a scenario where $M$ malicious clients ($M < N$) infiltrate the FL system to disturb or manipulate the central aggregation process by transmitting false local updates.
Because FL systems are decentralized, attackers cannot access updates from benign users and hence have a limited view of the entire data distribution.
We consider two different types of poisoning attacks.
One is untargeted attacks, in which attackers may send Gaussian noise to the central server or train the local model with randomly swapped labels. Such tampering can harm the global model's performance.
The other type is targeted attacks, in which attackers send model updates containing a backdoor with a carefully designed backdoor trigger.
This will cause the global model to incorrectly classify test samples under a specific target label. \looseness=-1

\cutparagraphup
\paragraph{Attack-tolerant central aggregation.} 
Most FL systems assume that all participants are benign and that their local updates are reliable.
This assumption leaves the system vulnerable to attacks that try to alter or manipulate updates for malicious purposes.
Attack-tolerant central aggregation methods have been proposed to mitigate the impact of malicious updates~\cite{fu2019attack,xie2018generalized,yin2018byzantine}.

Let $\mathcal{C}_m$ denote a set of malicious clients and $\mathcal{C}_{b}$ a set of benign clients, $\mathcal{C} = \mathcal{C}_m \cup \mathcal{C}_b$.
Then, the objective of attack-tolerant central aggregation is to design the aggregation function $g^*(\cdot)$, which can be defined as follows,
\begin{align}
\phi^{t+1} &= \phi^t + {{\sum_{i\in [1..N]} \mathbf{1}(i \in \mathcal{C}_{b}) \cdot \Delta_i^t} \over N-M} \nonumber \\
&= \phi^t + \sum_{i\in [1..N]} g^*(i) \cdot \Delta_i^t,
\label{Eq:optimal_aggregation}
\end{align}
where $\mathbf{1}(i \in \mathcal{C}_{b})$ is an indicator function that becomes one if client $i \in \mathcal{C}_b$ and zero if client $i \notin \mathcal{C}_b$. 
The term $|\mathcal{D}_i|$ in Eq.~\ref{eq:fedavg} is omitted here to prevent magnifying the effect of false updates by attackers with increased sizes of their datasets. \looseness=-1

%% file: contents/4_discovering_pattern.tex
\begin{figure*}[t!]
\centering
\begin{subfigure}[t]{0.32\textwidth}
\captionsetup{justification=centering}
       \centering\includegraphics[height=3.9cm]{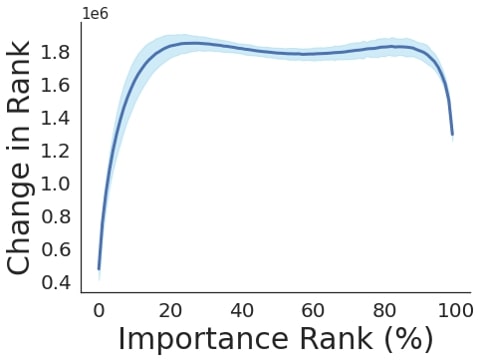}
      \caption{Changes of importance in benign clients}
      \label{fig:analysis1}
\end{subfigure}
\begin{subfigure}[t]{0.32\textwidth}
\captionsetup{justification=centering}
       \centering\includegraphics[height=3.9cm]{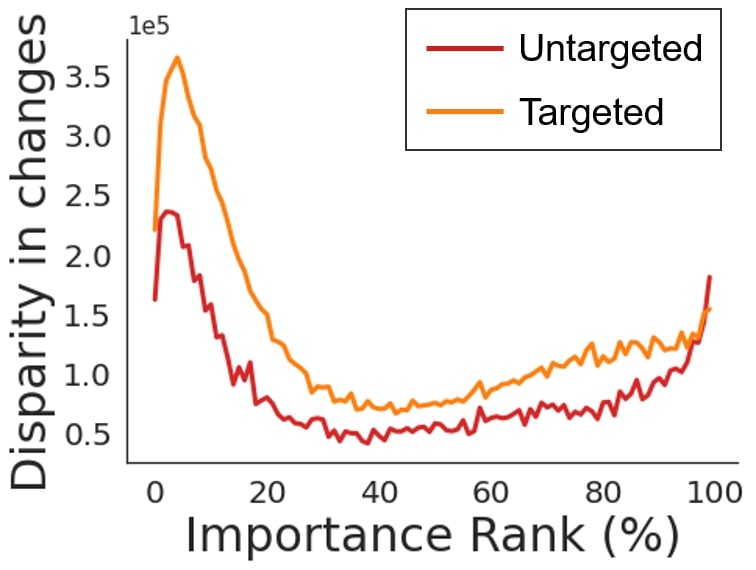}
      \caption{Disparity in changes by attacks}
      \label{fig:analysis2}
\end{subfigure}
\begin{subfigure}[t]{0.32\textwidth}
\captionsetup{justification=centering}
       \centering\includegraphics[height=3.9cm]{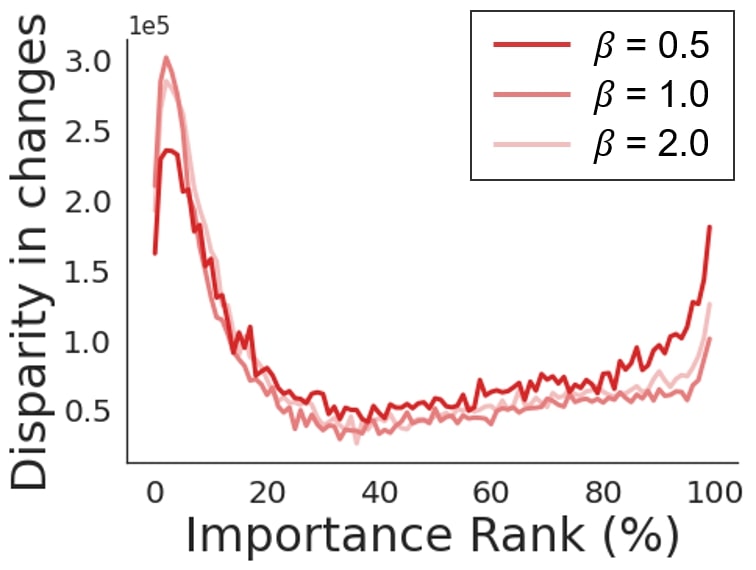}
      \caption{Disparity in changes with different $\beta$}
      \label{fig:analysis3}
\end{subfigure}
\caption{Analysis of importance-rank changes of parameters in federated learning: (a) Averaged change in importance-ranks of parameters in benign local models after one training round with the standard deviation area shaded. (b) Comparison of change patterns under two different poisoning attack scenarios, where the disparity is measured by the difference in importance-rank changes between benign and poisoned models after one training round. (c) The disparity in change patterns of the untargeted attack under varying data heterogeneity determined by $\beta$. \looseness=-1
}
\label{fig:critical_analysis}
\end{figure*}

\section{Critical Parameter Analysis}
\label{sec:parameter_analysis}

\noindent
Given the problem statement, our goal, as formulated in Eq.~\ref{Eq:optimal_aggregation}, is to determine which updates are malicious and neutralize their impact during the central aggregation process.
Prior studies used $L_2$ distance-based similarity, assuming that false updates are positioned far from benign updates in the Euclidean space~\cite{blanchard2017machine,fu2019attack}. Such an approach performs poorly in the non-IID setting~\cite{baruch2019little}, where benign updates become diverse enough to be separated from malicious updates. 
Motivated by a recent study that demonstrated parameters play diverse roles in model training~\cite{frankle2019lottery,lee2019snip,xia2021robust}, we adopt an alternative approach to examine parameter importance and identify common patterns among benign updates distinguishable from malicious updates. 
Our new defense strategy is robust under non-IID data distributions.

Let $\theta_i^t$ denote the model parameters of client $i$ at communication round $t$.
After the local training, the model update is defined as $\Delta_i^t = \theta_i^t - \phi^t$.
As originally used in~\cite{xia2021robust}, we evaluate the importance $p_i$ of the model parameters of client $i$ with the following equation: \looseness=-1
\begin{align}
p_i[n] = | \Delta_i[n] \cdot \theta_i[n]|,\label{eq:importance}
\end{align}
where the notation $[n]$ represents the $n$-th component value of a given vector.

The role of Eq.~\ref{eq:importance} is two-fold. First, the magnitude of the update provides information about the intensity of the learning signal imposed on each parameter for optimization~\cite{lee2019snip}.
Second, the magnitude of the weight represents how much the parameter contributes to the model's prediction~\cite{frankle2019lottery}.
By considering both the update and the weight, we can comprehensively assess the importance of each model parameter.
Specifically, when the value of $p_i[n]$ is large, the parameter is considered critical and can significantly impact the optimization process.
If $p_i[n]$ is small, the parameter is considered non-critical and is rarely used for training.

Given a federated learning system with multiple clients and parameter importance information of each local model, we conduct an analysis to answer the questions below.
\begin{itemize}
    \item \textbf{Q1.} \textit{Do benign local models exhibit similar patterns of changing parameter importance during training?} \looseness=-1

    \item \textbf{Q2.} \textit{Are there any differences in the change of parameter importance between the training of normal and malicious objectives?} \looseness=-1

    \item \textbf{Q3.} \textit{If any patterns are discovered, are they persistent across different non-IID settings and datasets?}
\end{itemize}


The first question asks about the common change pattern in the importance-ordering of local model parameters among benign clients.
To answer this question, we conducted multiple rounds of communication in the FL system using the CIFAR-10 dataset.
For each round $t>1$, the central server shares its global model, $\phi^t$, with clients. 
Clients then record the parameter importance of the shared global model $\phi^t$ via $p^t_{global}[n]=|\Delta^t[n] \cdot \phi^t[n]|$, where $\Delta^t[n] = \phi^t[n] - \phi^{t-1}[n]$ is the change of the global model made from the previous $t-1$ round. Note that $p^t_{global}$ is identical for all clients since they receive the same model.
The parameters were then ordered according to the global model's parameter importance $p^t_{global}$ (x-axis in Figure~\ref{fig:critical_analysis}).
After the local training, each client $i$ computes the model's importance again, expressed as $p^t_{i}$. We analyzed the changes in orderings between $p^t_{global}$ and $p^t_{i}$ for each round $t>1$, and the averaged results are displayed in Fig.~\ref{fig:critical_analysis}\footnote{Note that the scale of the y-axis in Fig.~\ref{fig:critical_analysis} lies within [0, 1.1E7], as we used the ResNet18 model with 1.1E7 trainable parameters.}. \looseness=-1


Figure~\ref{fig:analysis1} shows experimental results of importance-rank changes in benign clients. 
We can see that most rank changes concentrate on parameters of medium importance, whereas the top-importance parameters tend to remain stable and the bottom-importance parameters tend to change less in importance ranks. 
This finding suggests different roles for parameters in the model;
The top-importance parameters may be less susceptible to changes due to their significant role in shaping the model's predictions.
On the other hand, the bottom-importance parameters have only a small effect on the prediction, and hence they may be neglected, resulting in fewer importance-rank changes during the optimization process.
A similar observation is also made in~\cite{xia2021robust} on the role of model parameters.

The experiment was then repeated in the presence of a poisoning attack.
We prepared two models derived from the same global model: one was trained with a normal objective and the other with a malicious objective.
The disparity in importance-rank changes between the two models was then computed for both targeted and untargeted attack scenarios.
The results are shown in Figure~\ref{fig:analysis2}. We can see that the poisoned models for both attack scenarios cause greater perturbations in the top- and bottom-importance parameters.
This phenomenon may be explained by the fact that a poisoning attack seeks to alter the most critical parameters for disrupting model optimization and injecting malicious information by awakening the unused parameters (i.e., less important parameters) to cause overfitting to noise.

Finally, we examined if this phenomenon holds for various non-IID data settings and datasets.
The level of non-IIDness was adjusted by the beta hyper-parameter ($\beta$) in the Dirichlet distribution of clients. The experimental results are shown in 
Figure~\ref{fig:analysis3} and Figure~\ref{fig:appendix_critical_analysis} in the Appendix. They both confirm that the pattern persists across varying levels of non-IIDness (adjusted by the $\beta$ value) and multiple datasets.
These observations can be summarized to answer the initial questions as follows: \looseness=-1
\begin{itemize}
    \item \textbf{A1.} \textit{When it comes to importance-rank changes of parameters, benign local models in FL systems tend to have similar top and bottom parameters in terms of importance ranks.}

    \item \textbf{A2.} \textit{Poisoned local models in FL systems tend to have different sets of parameters with top and bottom importance compared with benign models, which can either degrade optimization or induce overfitting.} \looseness=-1

    \item \textbf{A3.} \textit{The above importance-rank change patterns of parameters persist for different levels of data heterogeneity and datasets.}
\end{itemize}

%% file: contents/5_model.tex
\section{Main Defense Approach: \model{}}

\noindent
We present an effective defense method against poisoning attacks, called \model{}.
Our key idea is to define a new model similarity metric through critical parameter analysis and measure the normality of each local update based on this similarity.
The model then attempts to filter out and reduce the impact of potentially malicious updates using attack-tolerant central aggregation.
We describe each procedure in detail. \looseness=-1

\cutparagraphup
\paragraph{Defining local model similarity.}
Given two local models and their parameter importance computed by Eq.~\ref{eq:importance}, we measure the local model similarity with two criteria: top/bottom-$k$ critical sets similarity and importance rank correlation.
First, we extract the indices of the top-$k$ and bottom-$k$ important parameters from each client (i.e., $\Theta_i^{\text{top}}$, $\Theta_i^{\text{bottom}}$ in client $i$) and compare them by calculating the Jaccard similarity between each pair of parameter sets.
Second, to further assess the similarity of the parameter importance pattern, we compute the Spearman correlation of the importance values between two models for both top-$k$ and bottom-$k$ parameter sets.
The correlation is calculated over the parameter sets that are common in the two models (i.e., $\Theta_{i\cap j}^{\text{top}}=\Theta_i^{\text{top}} \cap \Theta_j^{\text{top}}$, $\Theta_{i\cap j}^{\text{bottom}}=\Theta_i^{\text{bottom}} \cap \Theta_j^{\text{bottom}}$).
These criteria are derived from our observations that the benign local model tends to have similar sets of parameters with top and bottom importance, while poisoned models do not.
The similarity measure between local models $\theta_i$ and $\theta_j$ is defined as the following equation: \looseness=-1
\begin{align}
    \text{sim}(\theta_i, \theta_j) =  &J(\Theta_i^{\text{top}}, \Theta_j^{\text{top}}) + J(\Theta_i^{\text{bottom}}, \Theta_j^{\text{bottom}}) \nonumber \\
    &+ \rho(r_i(\Theta_{i\cap j}^{\text{top}}), r_j(\Theta_{i\cap j}^{\text{top}}))  \nonumber \\ &+ \rho(r_i(\Theta_{i\cap j}^{\text{bottom}}), r_j(\Theta_{i\cap j}^{\text{bottom}})), \label{eq:model_similarity}
\end{align}
where $J(\cdot, \cdot)$ denotes the Jaccard similarity and $\rho(\cdot, \cdot)$ denotes the Spearman correlation between two inputs, which is normalized to [0, 1] to align the scale. Here,  $r_i$ and $r_j$ represent the functions that map indices to their ranks in terms of parameter importance for clients $i$ and $j$, respectively. \looseness=-1

\input{contents/algo}

\cutparagraphup
\paragraph{Normality score for local model.}
Assuming that adversarial models would have dissimilar patterns of parameter importance from other benign models, we regard a model with low similarity to others as likely malicious.
Given the set of clients $\mathcal{C}^t$ participating in communication round $t$, normality score $\mathcal{N}(\theta_i^t)$ of the local model $\theta_i^t$ can be defined as follows: \looseness=-1
\begin{align}
    \mathcal{N}(\theta_i^t) = {1 \over {|\mathcal{C}^t|}} \sum_{j \in \mathcal{C}^t} \text{sim}(\theta_i^t, \theta_j^t). \label{eq:normality_score_old}
\end{align}

However, relying solely on similarities among local models is susceptible to a Sybil attack, where most clients selected at the beginning of the round are malicious by chance~\cite{fung2020limitations}.
In this scenario, the normality score for adversarial models can be overestimated, as their updates tend to be similar.
To enhance the stability of the defense, we also compare the local model with the global model $\phi^{t-1}$ from the previous $t-1$ round, resulting in the following normality score,
\begin{align}
    \mathcal{N}(\theta_i^t) = \text{sim}(\theta_i^t, \phi^{t-1}) + {1 \over {|\mathcal{C}^t|}} \sum_{j \in \mathcal{C}^t} \text{sim}(\theta_i^t, \theta_j^t).  \label{eq:similarity_final}
\end{align}

\input{table/table_1_comparison_targeted_attack}
\input{table/table_2_comparison_label_flipping_attack}

\cutparagraphup
\paragraph{Attack-tolerant central aggregation.}
We aggregate local updates through a weighted average, with the weight $\lambda^t_i$ determined by the normality score $\mathcal{N}(\theta_i^t)$.
This allows us to filter out the effect of likely malicious updates, while preserving the knowledge gained from likely benign clients' updates.
To convert normality scores into weights, we scale each score to the range from 0 to 1 with Min-Max normalization, i.e., $\mathcal{\Tilde{N}}(\theta_i^t) \leftarrow \text{Scale}(\mathcal{N}(\theta_i^t))$.
Following the literature~\cite{fung2020limitations}, we apply the inverse sigmoid function to a normalized score to enhance the differentiation of weight values and avoid over-penalization of low, non-zero similarity values on benign clients, resulting in the following weight, 
\begin{align}
    \lambda^t_i = \text{Clip}_{0\sim 1}(\ln {\mathcal{\Tilde{N}}(\theta_i^t) \over {1 - \mathcal{\Tilde{N}}(\theta_i^t)}} + 0.5).  \label{eq:lambda} 
\end{align}
where $\text{Clip}_{0\sim 1}(\cdot)$ denotes a function that rounds and clips any values exceeding the 0-1 range. 
Given the local update from client $i$ as $\Delta_i$, the global model at communication round $t$ is updated as follows:
\begin{align}
    \phi^{t+1} \leftarrow \phi^t + {1 \over {\sum_{i \in \mathcal{C}^t} \mathbf{1}(\lambda^t_i > 0)}} \sum_{i \in \mathcal{C}^t} \lambda^t_i \cdot \Delta^t_i, \label{eq:final}
\end{align}
where $\mathbf{1}(\lambda^t_i > 0)$ is an indicator function that produces one if $\lambda^t_i$ is larger than zero and zero otherwise.
The overall procedure of \model{} is described in the Algorithm~\ref{algo:overall}. \looseness=-1

%% file: contents/algo.tex
\begin{algorithm}[t!]
\DontPrintSemicolon
\SetAlgoLined
\SetNoFillComment 
\caption{Central aggregation process with \model{}}
\begin{flushleft}
\vspace{-3mm}
 \textbf{Input:} Global model weight $\phi^t$, global model weight from previous round $\phi^{t-1}$, a set of local clients $\mathcal{C}^t$ with their models $\theta_i^t$, and updates $\Delta_i^t$, given index $i$. \smallskip
 
    
\small{\tcp{Computing parameter importance}}
\For{each client $i \in \mathcal{C}^t$}
{
    $p_i^t[n] = | \Delta_i^t[n] \cdot \theta_i^t[n]|$  in Eq.~\ref{eq:importance} \\
    $\Theta_i^{\text{bottom}}, \Theta_i^{\text{top}} = \argsort(p_i^t)[:k],\ \   \argsort(p_i^t)[-k:]$  \\
}

\small{\tcp{Measuring normality score}}
\For{each client $i \in \mathcal{C}^t$}
{
    \For{each client $j \in \mathcal{C}^t, j\neq i$}
    {
        $\Theta_{i\cap j}^{\text{top}}=\Theta_i^{\text{top}} \cap \Theta_j^{\text{top}}$, \ \ $\Theta_{i\cap j}^{\text{bottom}}=\Theta_i^{\text{bottom}} \cap \Theta_j^{\text{bottom}}$ \\
        $s_{i,j} = \text{sim}(\theta_i^t, \theta_j^t)$ in Eq.~\ref{eq:model_similarity}
    }
    $\mathcal{N}(\theta_i^t)$ = $\text{sim}(\theta_i^t, \phi^{t-1}) + {1 \over {|\mathcal{C}^t|}} \sum_{j \in \mathcal{C}^t} s_{i,j}$ in Eq.~\ref{eq:similarity_final} \\
    $\mathcal{\Tilde{N}}(\theta_i^t)$ = Scale($\mathcal{N}(\theta_i^t)$) \\
    $\lambda_i^t = \text{Clip}_{0\sim 1}(\ln {\mathcal{\Tilde{N}}(\theta_i^t) \over {1 - \mathcal{\Tilde{N}}(\theta_i^t)}} + 0.5)$ in Eq.~\ref{eq:lambda} \\
}
\small{\tcp{Attack-tolerant update}}
$   \phi^{t+1} \leftarrow \phi^t + {1 \over {\sum_{i \in \mathcal{C}^t} \mathbf{1}(\lambda^t_i > 0)}} \sum_{i \in \mathcal{C}^t} \lambda^t_i \cdot \Delta^t_i$ in Eq.~\ref{eq:final}
\end{flushleft}
\vspace{-2mm}
\label{algo:overall}
\end{algorithm}

%% file: table/table_1_comparison_targeted_attack.tex
\begin{table*}[t!]
\setlength{\tabcolsep}{3.5pt}
\centering
\begin{subtable}[t]{.49\textwidth}
    \centering
    \resizebox{1\columnwidth}{!}{
    \begin{tabular}{l|cc|cc|cc}
    \toprule
    Method & \multicolumn{2}{c|}{CIFAR-10}  & \multicolumn{2}{c|}{SVHN} & \multicolumn{2}{c}{TinyImageNet}  \\ 
    ($\gamma_p = 0.5$) & ACC($\uparrow$)  & ASR($\downarrow$)  &  ACC & ASR  & ACC  &  ASR\\ \midrule
     No Defense            &  72.1     &  71.0 & 93.0 &22.2 & 39.5    & 96.6 \\
    Median               &	65.6 &  77.8 & 90.7 & 23.0 &  32.5 &  96.1 \\
    Trimmed Mean                &    70.1  &  51.4    & 92.2  & 20.9 & 39.3 & 97.2\\
    Multi Krum                   &  69.9      &  63.8   & 92.1 & 21.4 &  37.1&  74.6\\     
    FoolsGold              & 45.5   &   54.3  &  79.6 &  23.5 &  24.3  & 92.4\\
    Norm Bound            &  68.2     &  61.2   &  93.1 &  20.8 & 36.6  & 96.7 \\ 
    RFA                   &  \textbf{72.8}      &  56.4   &  92.3 & 20.8 &   37.1 & 93.9  \\     
    ResidualBase         & 70.6  &  59.9     &  93.1  &  21.1 &   \textbf{39.6}   &  96.9 \\
    \model{}             & 68.8  &   \textbf{21.9}    & \textbf{93.3}     &  \textbf{20.6}  &  30.1   &  \textbf{43.2} \\ \bottomrule
    \end{tabular}
    }
\end{subtable}
\hspace{1mm}
\begin{subtable}[t]{.49\textwidth}
    \centering
    \resizebox{1\columnwidth}{!}{
    \begin{tabular}{l|cc|cc|cc}
    \toprule
Method & \multicolumn{2}{c|}{CIFAR-10}  & \multicolumn{2}{c|}{SVHN} & \multicolumn{2}{c}{TinyImageNet}  \\ 
 ($\gamma_p = 0.8$)   & ACC($\uparrow$)  & ASR($\downarrow$)   &  ACC & ASR  & ACC  &  ASR\\ \midrule
 No Defense            &   69.3    &  50.9 &  92.5  & 22.0 & \textbf{38.8} & 96.1\\
Median               &	 62.4 & 70.6 & 90.0 &  23.6 &  31.5   & 96.2\\
Trimmed Mean                &  71.4    & 19.0  &  91.7  &  21.4  & 37.9 & 97.0\\
Multi Krum                   &  69.0      &   40.4   &   90.7 & 23.4 & 36.3  & 19.0 \\      
FoolsGold               &  49.1 &   46.8    &  69.8   & 32.3  & 28.5    & 69.1\\
Norm Bound              &    64.9    &  53.1   & 92.7  & 20.9 & 35.7  & 97.1 \\   
RFA                   &    70.1      &  44.8   & 91.8 & 22.1 &  36.3 & 11.4 \\     

ResidualBase         & 69.9  &   54.0    &   92.5  &  21.9 &  38.6  & 96.2\\
\model{}             &  \textbf{72.3} &  \textbf{12.5}     &  \textbf{93.1} & \textbf{20.8}  &  38.7  & \textbf{4.8} \\ \bottomrule
    \end{tabular}
    }
\end{subtable}
\caption{Comparison of defense performance over three datasets under targeted attack scenarios with different levels of pollution ratio $\gamma_p=0.5$, $0.8$. ACC and ASR refer to the final accuracy and the attack success rate, respectively. The symbol ($\uparrow$) indicates that a higher value is preferable, while ($\downarrow$) represents the opposite. The best results are marked bold.}
\label{Tab:targeted_attack}
\end{table*}

%% file: table/table_2_comparison_label_flipping_attack.tex
\begin{table*}[t!]
\centering
\begin{subtable}[t]{.49\textwidth}
\centering
\resizebox{0.95\columnwidth}{!}{\begin{tabular}{l|ccc}
\toprule
 Method ($\gamma_p=0.8$)  & CIFAR-10 & SVHN & TinyImageNet \\ \midrule
No Defense & 69.8 & 90.6 & 33.0 \\ 
Median  &  59.8  & 89.9  & 28.7 \\ 
Trimmed Mean  & 72.9 & 91.0 & 34.1 \\ 
Multi Krum & 72.7 & 92.6 & 35.9 \\ 
FoolsGold &  18.6 &  47.6  & 4.6 \\ 
Norm Bound    &  64.9        &   90.8  &  29.3 \\ 
RFA       &   72.6     &  92.7   &  36.5 \\     
ResidualBase & 73.6     & 92.1 & 36.0  \\
\model{} &  \textbf{74.9} &  \textbf{93.2} & \textbf{36.8}  \\\bottomrule
\end{tabular}}
\end{subtable}
\begin{subtable}[t]{.49\textwidth}
\centering
\resizebox{0.95\columnwidth}{!}{\begin{tabular}{l|ccc}
\toprule
 Method ($\gamma_p=1.0$)     & CIFAR-10 & SVHN & TinyImageNet \\ \midrule
No Defense & 63.8 & 86.1 & 24.4 \\ 
Median  & 56.8   & 89.6 & 21.2 \\ 
Trimmed Mean  & 66.2 & 87.9 & 27.2 \\ 
Multi Krum & 73.0 & 92.6 & 35.9 \\ 
FoolsGold &  24.9 &  41.9  & 1.3 \\ 
Norm Bound     &  63.5     &   86.6  &  24.1\\ 
RFA       &  71.5      &  92.4      &  \textbf{36.3} \\     
ResidualBase &  70.3    & 91.8  & 30.5  \\
\model{} & \textbf{74.4} & \textbf{93.2} & 34.9 \\\bottomrule
\end{tabular}}
\end{subtable}
\caption{Comparison of defense performance over three datasets under label flipping attack scenarios with different levels of pollution ratio $\gamma_p=0.8$, $1.0$. The best results are marked bold.} 
\label{Tab:untargeted_flipping}
\end{table*}

%% file: contents/6_result.tex
\section{Experiments}

\noindent
We evaluate the effectiveness of \model{} in defending against several attack scenarios over multiple datasets.
Component analyses are conducted to confirm the contribution of each component to robustness under varying simulation hyper-parameters.  \looseness=-1

\subsection{Defense Performance Evaluation}
\paragraph{Data.}
Three benchmark datasets on image classification tasks are utilized in our experiment: (1) CIFAR-10~\cite{krizhevsky2009learning} includes 60,000 samples of 32x32 pixels with 10 classes; (2) SVHN~\cite{netzer2011reading} includes 73,257 training samples and 26,032 test samples of 32x32 sized digits; (3) TinyImageNet~\cite{le2015tiny} contains 100,000 samples from 200 classes.

In our experiments, the non-IID property of federated learning in the three datasets is simulated using the Dirichlet distribution, following previous works~\cite{han2022fedx,li2021model}.
The Dirichlet distribution can be denoted as $Dir(N,\beta)$, where $N$ is the total number of clients and $\beta$ refers to the parameter that adjusts the level of heterogeneity in the decentralized data distributions.
A lower value of $\beta$ results in greater non-IIDness.
We set $N$ and $\beta$ to 20 and 0.5 as default values, respectively. \looseness=-1

\cutparagraphup
\paragraph{Implementation details.}
We set the number of communication rounds to 100, with one epoch of local training per round.
Following the literature~\cite{han2022fedx,li2021model}, we use ResNet18 as the default backbone network.
The SGD optimizer is employed. 
The learning rate, momentum, and weight decay parameter for the optimizer are set to 0.01, 0.9, and 1e-5.
The batch size is set to 64.
The hyper-parameter $k$ for top and bottom-$k$ parameter sets is set to 0.01 (1\%).
To simulate a more realistic federated setting, half of the clients (i.e., $N/2$) are randomly chosen in each round of training.
Data augmentation techniques such as random crop, horizontal flip, and color jitter are applied during the local training. In the case of the targeted attack, we follow the original literature~\cite{gu2017badnets} and generate a noise input pattern called backdoor.
The size of the backdoor is set to 5$\times$5, and its location is in the bottom-right corner of the images.
For the untargeted Gaussian attack, we set the standard deviation of the Gaussian noise to 0.05. \looseness=-1

\input{table/table_3_comparison_gaussian_attack}

\cutparagraphup
\paragraph{Baselines.}
A total of eight baselines are compared:
(1) \textsf{No Defense} represents the classical FedAvg algorithm without any consideration of attack scenarios;
(2) \textsf{Median} and (3) \textsf{Trimmed Mean}~\cite{xie2018generalized,yin2018byzantine} utilize the outlier-resistant statistics, mean and trimmed mean of local updates, for aggregation;
(4) \textsf{Multi Krum}~\cite{blanchard2017machine} iteratively selects a likely-benign local update with the lowest Euclidean distance from other updates;
(5) \textsf{FoolsGold}~\cite{fung2020limitations} identifies grouped actions of attacks by inspecting similarity among local updates;
(6) \textsf{Norm Bound}~\cite{sun2019can} filters out the updates whose norm is above a predefined threshold;
(7) \textsf{RFA}~\cite{pillutla2022robust} applies the geometric median operation for robust aggregation;
(8) \textsf{Residual Base}~\cite{fu2019attack} introduces a repeated median estimator to compute the confidence of each update.  \looseness=-1

For all baselines, we followed the original implementations and hyper-parameter settings. The confidence interval and clipping threshold in the ResidualBase algorithm are set to 2.0 and 0.05, respectively. 
In RFA, we set the smoothing parameter to 1e-6 and the maximum number of Weiszfeld iterations to 100. \looseness=-1

\input{table/table_4_comparison_summaries}

\cutparagraphup
\paragraph{Evaluation.}
All methods are assessed under the same experimental settings (e.g., $\beta$, the number of clients, communication rounds, and epochs for local training).
Given a total of $N$ clients, we set 20\% of the clients to play an adversarial role as default.
Three attack scenarios are evaluated, one for targeted and two for untargeted attacks.
The targeted attack injects a crafted backdoor trigger pattern into some training images and changes their labels to the target class to manipulate the model training.
The untargeted attacks consist of the label flipping attack, which randomly alters the labels to generate false update signals~\cite{xiao2012adversarial}, and the Gaussian noise attack, which sends Gaussian noise as an update~\cite{fang2020local}.
For both the targeted and label flipping attacks, experiments were conducted with two different levels of pollution ratio ($\gamma_p$), representing the fraction of poisoned samples added to the dataset.
In the targeted attack experiments, we use a pollution ratio of 0.5 and 0.8, while a pollution ratio of 0.8 and 1.0 is used in the label flipping attack experiments.
As an evaluation metric, the final accuracy on the test set is reported for the untargeted attack scenarios, while both the attack success rate and the final accuracy are reported for the targeted attack scenario. 
All measures are calculated by averaging the last ten rounds of results.
\looseness=-1

\begin{figure*}[t!]
\centering
\begin{subfigure}[t]{0.32\textwidth}
\captionsetup{justification=centering}
       \centering\includegraphics[height=3.9cm]{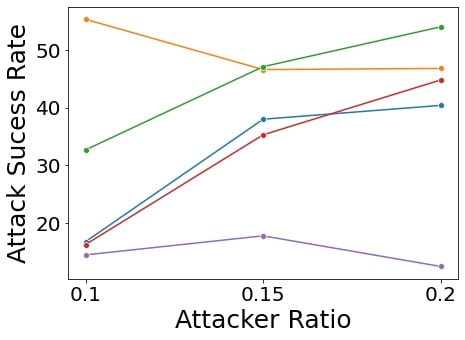}
      \caption{Effect of the ratio of attackers}
      \label{fig:test1}
\end{subfigure}
\begin{subfigure}[t]{0.32\textwidth}
\captionsetup{justification=centering}
       \centering\includegraphics[height=3.9cm]{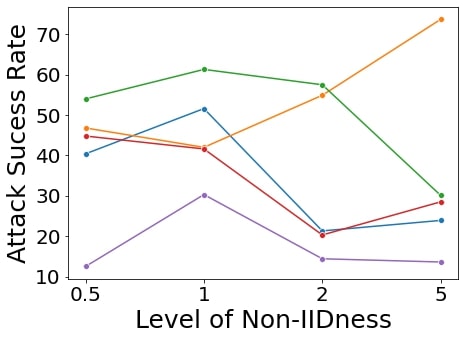}
      \caption{Effect of the level of non-IIDness}
      \label{fig:test2}
\end{subfigure}
\begin{subfigure}[t]{0.32\textwidth}
\captionsetup{justification=centering}
       \centering\includegraphics[height=3.9cm]{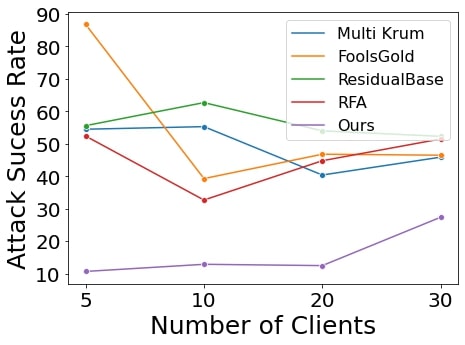}
      \caption{Effect of the number of clients}
      \label{fig:test3}
\end{subfigure}
\caption{Robustness test results against targeted attacks on CIFAR-10 with varying experimental settings. The results demonstrate that \model{} consistently achieves the best defense performance (lowest ASR) compared with the baselines. \looseness=-1
}
\label{fig:robustness_test}
\end{figure*}

\input{table/table_5_ablation_study}

\cutparagraphup
\paragraph{Results.}
Tables~\ref{Tab:targeted_attack}-\ref{Tab:result_summary} present the evaluation results and their summaries for different attack scenarios.
\model{} shows the best or comparable classification accuracy and attack success rate against other defense strategies over all datasets.
Our method consistently performs satisfactorily against all types of attacks, whereas some baselines may struggle against specific attacks (e.g., ResidualBase in the targeted attack scenario).
Notably, \model{} reduces the success rate of targeted attacks by a factor of 2 to 4 compared to other baselines on the CIFAR-10 and TinyImageNet datasets.
These results highlight the effectiveness of our method in providing robustness for FL systems. \looseness=-1

\subsection{Component Analysis}
\paragraph{Ablation study.} 
We conduct an ablation study to evaluate the contribution of each component in our full model.
The following variations are compared: (1) \textsf{without topk} only considers and compares parameters of bottom-$k$ importance to compute the normality score of models, while (2) \textsf{without bottomk} is vice-versa; (3) \textsf{without global} omits the similarity term in the normality score between the local model and the global model from the previous round (Eq.~\ref{eq:normality_score_old}); (4) \textsf{without local} only utilizes global model similarity for the normality measure (i.e., $\mathcal{N}(\theta^t_i) = \text{sim}(\theta_i^t, \phi^{t-1})$).

Table~\ref{Tab: Ablation} shows that the full model with all components performs the best against both targeted and untargeted attacks (i.e., label flipping attacks) among all variations, which implies that each component plays an important role in detecting malicious updates.
Interestingly, without considering the bottom-$k$ important parameters, the ablation study showed the greatest decrease in defense performance among all the ablations.
These results support our hypothesis that poisoning attacks cause a local model to overfit maliciousness by utilizing unused parameters.
Therefore, simply focusing on the bottom-$k$ important parameters is also effective in detecting adversarial clients.

\cutparagraphup
\paragraph{Robustness test.}
Next, we conduct experiments in settings with varying key experimental parameters to assess the robustness of our approach.
These include (a) the number of malicious clients $|\mathcal{C}_m|$, (b) the total number of participating clients $N$, and (c) the degree of non-IIDness, controlled by $\beta$ in the Dirichlet distribution.

The performance comparison between \model{} and the baselines on the CIFAR-10 dataset is shown in Figure~\ref{fig:robustness_test}. 
Only the results for the targeted attack scenario are reported due to the space limitation. More results can be found in the appendix. 
We can see that, under various experimental settings, \model{} consistently demonstrates superior defense performance. \looseness=-1

\input{table/table_6_hyperparameter_analysis}

\begin{figure*}[t!]
\centerline{\includegraphics[width=0.87\linewidth]{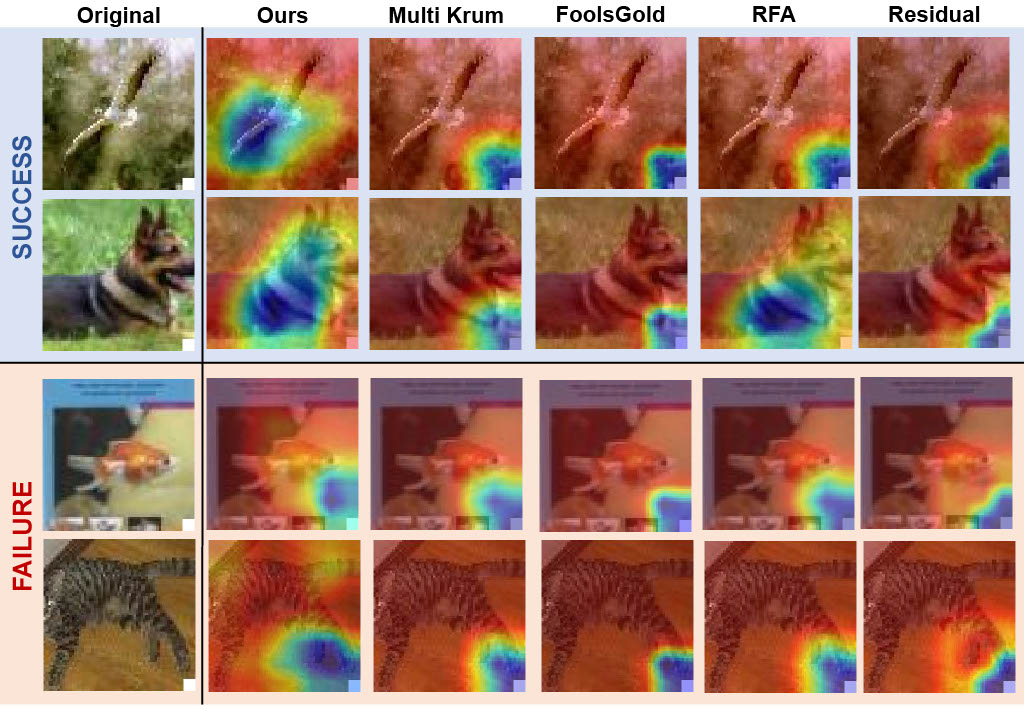}}
\caption{Qualitative analysis under a targeted attack scenario over TinyImagenet, where the highlighted part visualizes how the model
recognizes class characteristics based on the Grad-CAM algorithm.}
\label{fig:qualitative_appendix}
\end{figure*}

\cutparagraphup
\paragraph{Hyper-parameter analysis.}
We investigate the effect of hyper-parameter $k$ on defense performance.
Hyper-parameter $k$ determines the proportion of model parameters selected to create the parameter sets $\Theta_i^{\text{top}}$ and $\Theta_i^{\text{bottom}}$ (i.e., the top-$k$ and bottom-$k$ most important parameters) for each client $i$.
The smaller $k$, the fewer parameters are compared to compute the normality score of the model.

The results for various values of $k$ are presented in Table~\ref{Tab: topk}.
Our method demonstrates satisfactory results for most measures under both targeted and untargeted attack (i.e., label flipping attack) scenarios when $k$ is within a reasonable range of 1-2\%.
However, setting $k$ to a value too small or too large significantly decreases the performance.
This is because the normality measure with a small $k$ may not have enough evidence to distinguish malicious updates, while the measure with a large $k$ can be disturbed by the importance changes caused by data heterogeneity. \looseness=-1

\cutparagraphup
\paragraph{Qualitative Analysis}
We also perform a qualitative analysis to assess how effectively \model{} can filter out malicious knowledge during training under targeted attack scenarios.
Figure~\ref{fig:qualitative_appendix} compares the performance of different defense strategies in interpreting class characteristics after training.
To evaluate each model's interpretation, we corrupted test set images with a small patch of noise used by attackers and used the Grad-CAM algorithm~\cite{selvaraju2017grad} to visualize the model's attention for each input.
Blue-framed images represent success cases randomly sampled from the dataset, while red-framed images represent failure cases.
Our method tends to extract key features from the image compared to other cases where the model is contaminated by malicious knowledge and only focuses on the injected noise patch.
Even in failure cases, our approach gives attention to other visual traits along with the noise, demonstrating its robustness against attacks. \looseness=-1

%% file: table/table_3_comparison_gaussian_attack.tex
\begin{table}[t!]
\centering
\scalebox{0.94}{
\begin{tabular}{l|ccc}
\toprule
Method       & CIFAR-10 & SVHN & TinyImageNet
 \\ \midrule
No Defense & 32.7 & 47.8 &  2.1 \\ 
Median  & 67.8   &  91.5 &  28.8\\ 
Trimmed Mean  & 55.6  & 72.5 &  12.1 \\ 
Multi Krum & 52.8 & 68.4 &  15.0 \\ 
FoolsGold & 13.9 &  6.7  &  0.5 \\ 
Norm Bound    &   28.2     & 42.9    & 1.2 \\ 
RFA       &   72.0    & 92.2    & 35.8 \\     
ResidualBase &   74.6   & \textbf{93.7}  & \textbf{37.0}  \\
\model{} & \textbf{74.8}  & 93.6 &36.1  \\\bottomrule
\end{tabular}}
 \label{Tab:untargeted_gaussian}
\caption{
Accuracy (\%) under the Gaussian noise attack over three datasets. The best results are marked bold.}
\end{table}

%% file: table/table_4_comparison_summaries.tex
\begin{table}[!t]
\setlength{\tabcolsep}{3.5pt}
\centering
\scalebox{0.9}{\begin{tabular}{l|cc|c|c|c}
\toprule
\multirow{2}{*}{Setup} & \multicolumn{2}{c|}{Targeted} & Label flipping & Gaussian & \multirow{2}{*}{Total} \\  
& ACC    &  ASR  & ACC & ACC & 
\\ \midrule
No Defense & 2.8 & 6.2 & 6.5 & 7.0 & 5.6 \\
Median & 7.8 & 7.5 & 7.5 & 4.0 & 6.7 \\
Trimmed Mean & 4.2 & 4.7 & 4.8 & 5.3 & 4.8 \\
Multi Krum & 5.7 & 4.7 & 2.8 & 5.7 & 4.7 \\
FoolsGold & 9.0 & 5.5 & 9.0 & 9.0 & 8.1 \\
Norm Bound & 5.2 & 5.5 & 6.8 & 8.0 & 6.4 \\
RFA & 3.8 & 3.7 & 2.7 & 3.0 & 3.3 \\
ResidualBase & 2.7 & 6.0 & 3.5 & 1.3 & 3.4 \\
\model{} & 3.2 & 1.0 & 1.3 & 1.7 & \textbf{1.8} \\
\bottomrule 
\end{tabular}}
\caption{Performance comparison summaries among defense strategies. Averaged rank for each evaluation metric under different attack scenarios, including both untargeted and targeted attacks, is reported. Our \model{} presents superb defense performance. \looseness=-1}
\label{Tab:result_summary}
\end{table}

%% file: table/table_5_ablation_study.tex
\begin{table}[!t]
\centering
\scalebox{0.97}{\begin{tabular}{l|cc|c}
\toprule
\multirow{2}{*}{Setup} & \multicolumn{2}{c|}{Targeted} & Untargeted \\  
& ACC    &  ASR  & ACC
\\ \midrule
All components     & \textbf{72.3}  & \textbf{12.5} & \textbf{74.9}\\
without topk             &      70.4  & 18.9            & 74.0 \\ 
without bottomk   &       60.5  & 36.8  & 67.6 \\
without global            &     68.8      & 24.1 & 74.8\\
without local            &     65.0      & 20.2 & 72.4 \\
\bottomrule 
\end{tabular}}
\caption{Ablation study results of \model{} on CIFAR-10. The best results are marked bold. Our method with full components reports the best defense performance against both targeted and untargeted attacks. \looseness=-1}
\label{Tab: Ablation}
\end{table}

%% file: table/table_6_hyperparameter_analysis.tex
\begin{table}[!t]
\centering
\scalebox{1}{\begin{tabular}{l|cc|c}
\toprule
\multirow{2}{*}{Top/bottom-$k$ ratio} & \multicolumn{2}{c|}{Targeted} & Untargeted \\  
& ACC    &  ASR  & ACC
\\ \midrule
$k$ = 0.005 (0.5\%)    & 61.0  & 63.4 & 71.4 \\
$k$ = 0.01 (1\%)    &    72.3  & 12.5 & 74.9\\ 
$k$ = 0.02 (2\%)     &   67.7      & 15.6  & 74.0 \\
$k$ = 0.05 (5\%)     &    60.2     &   51.6 & 74.3 \\
\bottomrule 
\end{tabular}}
\caption{Hyper-parameter analysis under both targeted and untargeted attacks on CIFAR-10 with different values of $k$.
}
\label{Tab: topk}
\end{table}

%% file: contents/7_conclusion.tex
\section{Conclusion}

\noindent
We presented \model{}, a defense strategy against poisoning attacks in federated learning systems.
Our method is based on the observation that benign local models tend to have similar sets of important parameters, while adversarial models do not.
To distinguish malicious updates, we propose a new normality measure that considers the pattern of important parameters in local models.
Then, we aggregate local updates via a weighted average, where the weight of a local update is determined by its normality score.
Extensive experiments with both targeted and untargeted attack scenarios on multiple datasets demonstrate the effectiveness of \model{} in defending against poisoning attacks.
Our work contributes to the ongoing efforts on attack-tolerant federated learning and provides new insights for future research.

\small{\paragraph{Acknowledgements.}
This research was supported by the Institute for Basic Science (IBS-R029-C2). Sungwon Han, Sungwon Park, and Meeyoung Cha were supported by the National Research Foundation of Korea (NRF) grant (RS-2022-00165347). Sundong Kim also received the NRF grant funded by the Ministry of Science and ICT (RS-2023-00240062).
}

\newpage

%% file: contents/8_appendix.tex
\appendix

\begin{figure*}[ht!]
\centering
\begin{subfigure}[t]{0.32\textwidth}
\captionsetup{justification=centering}
       \centering\includegraphics[height=3.9cm]{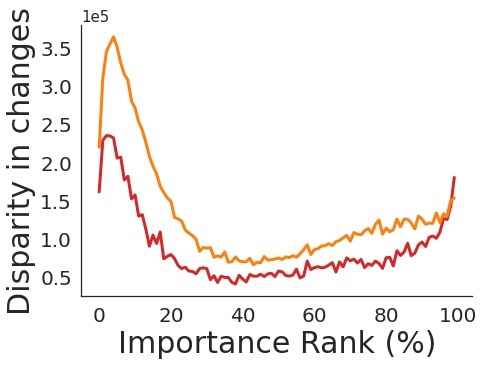}
      \caption{CIFAR-10}
      \label{fig:appendix_analysis1}
\end{subfigure}
\begin{subfigure}[t]{0.32\textwidth}
\captionsetup{justification=centering}
       \centering\includegraphics[height=3.9cm]{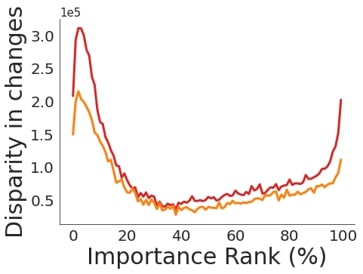}
      \caption{SVHN}
      \label{fig:appendix_analysis2}
\end{subfigure}
\begin{subfigure}[t]{0.32\textwidth}
\captionsetup{justification=centering}
       \centering\includegraphics[height=3.9cm]{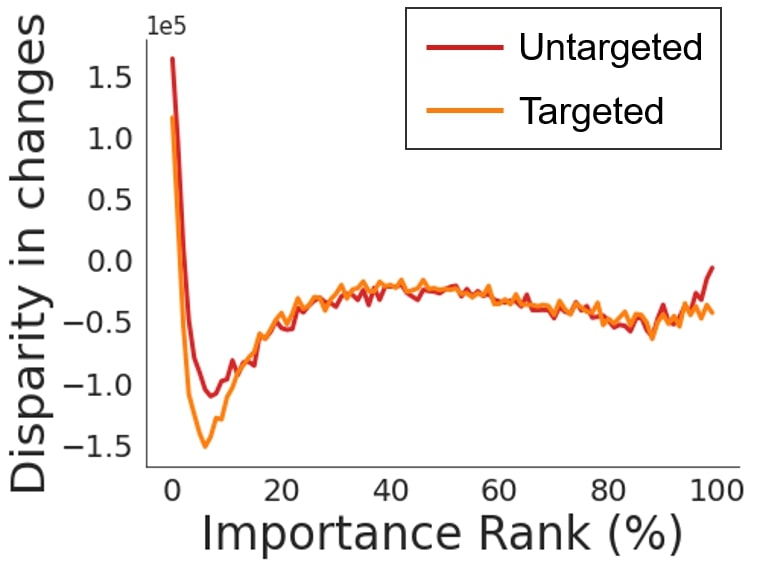}
      \caption{TinyImageNet}
      \label{fig:appendix_analysis3}
\end{subfigure}
\caption{Comparison of change patterns over three datasets under two different poisoning attack scenarios, untargeted and targeted attack, where the disparity is measured by the difference in changes of importance rank between benign and poisoned models after one training round. \looseness=-1
}
\label{fig:appendix_critical_analysis}
\end{figure*}

\section{Appendix}
\cutsubsectionup
\subsection{Release \& Implementation details}

\noindent
We adopt ResNet18 as the default backbone architecture, building upon prior research in federated learning~\cite{li2021model,han2022fedx}.
In the case of the targeted attack, we follow the original literature~\cite{gu2017badnets} and generate a noise input pattern called a backdoor.
The size of the backdoor is set to 5$\times$5, and its location is in the bottom-right corner of the images.
For the untargeted Gaussian attack, we set the standard deviation of the Gaussian noise to 0.05.

We follow the original works' implementations and hyper-parameter settings to reproduce all baselines.
For Multi-Krum and Norm Bounding algorithms, we assume the central server already knows the upper bound of attacker numbers when deciding on hyper-parameters.
The confidence interval and clipping threshold in the ResidualBase algorithm are set to 2.0 and 0.05, respectively. 
We calculate the geometric mean for RFA by setting the smoothing parameter to 1e-6 and the maximum number of Weiszfeld iterations to 100.
More details on implementations are at \url{https://github.com/Sungwon-Han/FEDCPA}.

\cutsubsectionup
\subsection{Time Complexity Analysis}

\noindent
For all experiments, we utilized four A100 GPUs.
Table~\ref{Tab:appendix_complexity} compares the time costs in seconds of every defense strategy per each round of training.
Note that \model{} is not a huge burden and only took 10\% more processing time than the classical FedAvg algorithm (i.e., No Defense).

\input{table/table_appendix_computational_complexity}

\cutsubsectionup
\subsection{Extra Results on Critical Parameter Analysis}

\noindent
In Section~\ref{sec:parameter_analysis}, we have shown that benign and poisoned local models exhibit distinct patterns in terms of parameter importance, with the poisoned model causing more significant disruptions to the top and bottom parameters.
We conducted the same analysis across different datasets to validate our observation.
The results of our analysis are presented in Figure~\ref{fig:appendix_critical_analysis}, which compares the change patterns in importance rank between benign and poisoned models under two different attack scenarios, untargeted and targeted attacks.
For the untargeted attack scenario, we used the label flipping attack method.
After one training round, We measure the disparity in importance rank between benign and poisoned models.
The results demonstrate that our observation remains consistent across the various datasets. \looseness=-1

\subsection{Extra Results on Robustness Tests}

\noindent
We evaluate the robustness of \model{} through experiments conducted under different settings, varying key simulation parameters such as (a) the number of malicious clients $|\mathcal{C}_m|$, (b) the total number of participating clients $N$, and (c) the degree of non-IIDness, controlled by the $\beta$ parameter in the Dirichlet distribution. 
A lower $\beta$ value results in a higher level of non-IIDness. \looseness=-1

\begin{figure*}[ht!]
\centering
\begin{subfigure}[t]{0.32\textwidth}
\captionsetup{justification=centering}
       \centering\includegraphics[height=3.9cm]{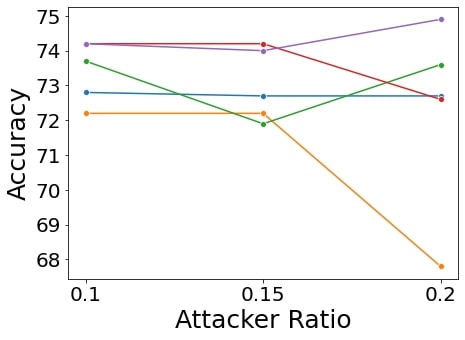}
      \caption{Effect of the ratio of attackers}
      \label{fig:test1_appendix}
\end{subfigure}
\begin{subfigure}[t]{0.32\textwidth}
\captionsetup{justification=centering}
       \centering\includegraphics[height=3.9cm]{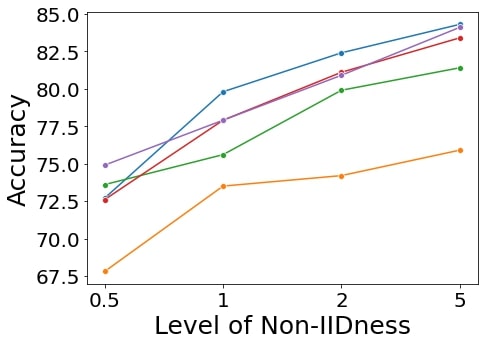}
      \caption{Effect of the level of non-IIDness}
      \label{fig:test2_appendix}
\end{subfigure}
\begin{subfigure}[t]{0.32\textwidth}
\captionsetup{justification=centering}
       \centering\includegraphics[height=3.9cm]{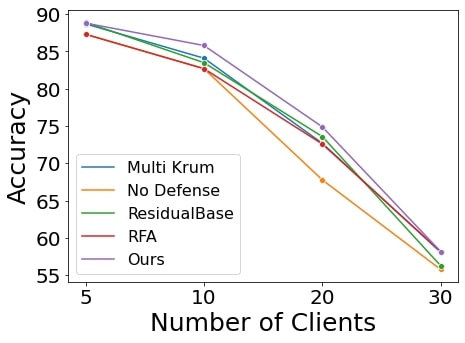}
      \caption{Effect of the number of clients}
      \label{fig:test3_appendix}
\end{subfigure}
\caption{Robustness test results under label flipping attack across different simulation hyper-parameters: (a) the attacker ratio, (b) the level of non-IIDness, and (c) the number of clients over the CIFAR-10 dataset. \looseness=-1
}
\label{fig:robustness_test_appendix}
\end{figure*}

This section presents additional comparison results among different defense strategies under an untargeted attack scenario (i.e., label flipping attack) on the CIFAR-10 dataset.
Results presented in Figure~\ref{fig:robustness_test_appendix} show that \model{} consistently performs comparably well despite variations in simulation parameters. \looseness=-1

\subsection{Full Results on Performance Evaluation}

\noindent
Table~\ref{Tab:targeted_attack_appendix}-\ref{Tab:untargeted_gaussian_appendix} shows the complete evaluation results on defense performance over three datasets under various poisoning attack scenarios: targeted attack with $\gamma_p=0.5,\ 0.8$, untargeted label flipping attack with $\gamma_p=0.8,\ 1.0$, and untargeted Gaussian attack. 
The results are obtained by averaging over the last ten rounds and are reported with mean and standard deviation values. 
\looseness=-1

\input{table/table_appendix_full_results}

%% file: table/table_appendix_computational_complexity.tex
\begin{table}[h!]
\centering
\scalebox{0.94}{
\begin{tabular}{l|ccc}
\toprule
Method       & Time costs in seconds
 \\ \midrule
No Defense & 87 \\
Median & 86 \\
Trimmed Mean & 87 \\
Multi Krum & 87 \\
FoolsGold & 90 \\
Norm Bound & 86 \\
RFA & 91 \\
ResidualBase & 185 \\
\model{} & 96 \\ \bottomrule
\end{tabular}}
\caption{Comparison on time complexity among defense strategies against poisoning attacks. The CIFAR-10 dataset is used for the analysis.}
\label{Tab:appendix_complexity}
\end{table}

%% file: table/table_appendix_full_results.tex
\begin{table*}[t!]
\centering
\resizebox{1.9\columnwidth}{!}{
\begin{tabular}{l|cc|cc|cc}
\toprule
Method & \multicolumn{2}{c|}{CIFAR-10}  & \multicolumn{2}{c|}{SVHN} & \multicolumn{2}{c}{TinyImageNet}  \\ 
($\gamma_p = 0.5$) & ACC  & ASR  &  ACC & ASR  & ACC  &  ASR\\ \midrule
No Defense & 72.1$\pm$3.07 & 71.0$\pm$0.61 & 93.0$\pm$0.48 & 22.2$\pm$12.02 & 39.5$\pm$2.71 & 96.6$\pm$0.41 \\
Median & 65.6$\pm$3.54 & 77.8$\pm$1.09 & 90.7$\pm$0.44 & 23.0$\pm$7.02 & 32.5$\pm$3.43 & 96.1$\pm$0.58 \\
Trimmed Mean & 70.1$\pm$3.15 & 51.4$\pm$0.82 & 92.2$\pm$0.57 & 20.9$\pm$20.66 & 39.3$\pm$1.13 & 97.2$\pm$0.26 \\
Multi Krum & 69.9$\pm$0.89 & 63.8$\pm$1.24 & 92.1$\pm$0.82 & 21.4$\pm$10.88 & 37.1$\pm$2.85 & 74.6$\pm$6.93 \\
FoolsGold & 45.5$\pm$12.24 & 54.3$\pm$18.25 & 79.6$\pm$5.32 & 23.5$\pm$36.78 & 24.3$\pm$7.37 & 92.4$\pm$14.82 \\
Norm Bound & 68.2$\pm$4.12 & 61.2$\pm$25.00 & 93.1$\pm$0.69 & 20.8$\pm$0.91 & 36.6$\pm$0.38 & 96.7$\pm$0.69 \\
RFA & 72.8$\pm$3.09 & 56.4$\pm$13.52 & 92.3$\pm$1.09 & 20.8$\pm$1.57 & 37.1$\pm$0.48 & 93.9$\pm$0.63 \\
ResidualBase & 70.6$\pm$3.12 & 59.9$\pm$0.61 & 93.1$\pm$0.34 & 21.1$\pm$15.45 & 39.6$\pm$1.27 & 96.9$\pm$0.19 \\
\model{} & 68.8$\pm$3.74 & 21.9$\pm$0.73 & 93.3$\pm$9.36 & 20.6$\pm$2.69 & 30.1$\pm$1.51 & 43.2$\pm$44.66 \\\bottomrule
\end{tabular}
}
\caption{Comparison of defense performance over three datasets under targeted attack scenarios with pollution ratio $\gamma_p=0.5$. Mean and standard deviation over ten last rounds are reported.}
\label{Tab:targeted_attack_appendix}
\end{table*}

\begin{table*}[t!]
\centering
\resizebox{1.9\columnwidth}{!}{
\begin{tabular}{l|cc|cc|cc}
\toprule
Method & \multicolumn{2}{c|}{CIFAR-10}  & \multicolumn{2}{c|}{SVHN} & \multicolumn{2}{c}{TinyImageNet}  \\ 
($\gamma_p = 0.8$) & ACC  & ASR  &  ACC & ASR  & ACC  &  ASR\\ \midrule
No Defense & 69.3$\pm$3.74 & 50.9$\pm$25.09 & 92.5$\pm$0.93 & 22.0$\pm$2.21 & 38.8$\pm$1.12 & 96.1$\pm$1.34 \\
Median & 62.4$\pm$3.32 & 70.6$\pm$16.51 & 90.0$\pm$1.53 & 23.6$\pm$3.39 & 31.5$\pm$0.99 & 96.2$\pm$0.59 \\
Trimmed Mean & 71.4$\pm$2.77 & 19.0$\pm$10.29 & 91.7$\pm$1.25 & 21.4$\pm$1.78 & 37.9$\pm$1.12 & 97.0$\pm$0.81 \\
Multi Krum & 69.0$\pm$2.21 & 40.4$\pm$21.85 & 90.7$\pm$2.33 & 23.4$\pm$4.06 & 36.3$\pm$1.78 & 19.0$\pm$13.91 \\
FoolsGold & 49.1$\pm$9.46 & 46.8$\pm$34.83 & 69.8$\pm$24.56 & 32.3$\pm$27.72 & 28.5$\pm$4.27 & 69.1$\pm$43.20 \\
Norm Bound & 64.9$\pm$4.28 & 53.1$\pm$30.29 & 92.7$\pm$1.31 & 20.9$\pm$1.42 & 35.7$\pm$1.00 & 97.1$\pm$0.83 \\
RFA & 70.1$\pm$3.37 & 44.8$\pm$21.58 & 91.8$\pm$1.44 & 22.1$\pm$2.01 & 36.3$\pm$1.05 & 11.4$\pm$5.80 \\
ResidualBase & 69.9$\pm$3.59 & 54.0$\pm$27.50 & 92.5$\pm$0.81 & 21.9$\pm$2.34 & 38.6$\pm$0.47 & 96.2$\pm$0.81 \\
\model{} & 72.3$\pm$0.88 & 12.5$\pm$1.02 & 93.1$\pm$1.02 & 20.8$\pm$1.35 & 38.7$\pm$0.63 & 4.8$\pm$1.40 \\ \bottomrule
\end{tabular}
}
\caption{Comparison of defense performance over three datasets under targeted attack scenarios with pollution ratio $\gamma_p=0.8$. Mean and standard deviation over ten last rounds are reported.}
\label{Tab:targeted_attack_appendix2}
\end{table*}

\begin{table*}[t!]
\centering
\centering
\resizebox{1.3\columnwidth}{!}
{\begin{tabular}{l|ccc}
\toprule
 Method ($\gamma_p=0.8$)  & CIFAR-10 & SVHN & TinyImageNet \\ \midrule
No Defense & 69.8$\pm$3.49 & 90.6$\pm$1.80 & 33.0$\pm$4.76 \\
Median & 59.8$\pm$3.16 & 89.9$\pm$1.55 & 28.7$\pm$4.73 \\
Trimmed Mean & 72.9$\pm$3.47 & 91.0$\pm$1.49 & 34.1$\pm$3.73 \\
Multi Krum & 72.7$\pm$3.61 & 92.6$\pm$0.99 & 35.9$\pm$2.22 \\
FoolsGold & 18.6$\pm$7.53 & 47.6$\pm$19.76 & 4.6$\pm$3.36 \\
Norm Bound & 64.9$\pm$4.19 & 90.8$\pm$2.06 & 29.3$\pm$5.18 \\
RFA & 72.6$\pm$2.31 & 92.7$\pm$0.96 & 36.5$\pm$0.78 \\
ResidualBase & 73.6$\pm$3.40 & 92.1$\pm$1.03 & 36.0$\pm$3.38 \\
\model{} & 74.9$\pm$3.30 & 93.2$\pm$0.72 & 36.8$\pm$1.53 \\\bottomrule
\end{tabular}}
\caption{Comparison of defense performance over three datasets under label flipping attack scenarios with pollution ratio $\gamma_p=0.8$. Mean and standard deviation over ten last rounds are reported.} 
\label{Tab:untargeted_flipping_appendix}
\end{table*}

\begin{table*}[t!]
\centering
\centering
\resizebox{1.3\columnwidth}{!}
{\begin{tabular}{l|ccc}
\toprule
 Method ($\gamma_p=1.0$)  & CIFAR-10 & SVHN & TinyImageNet \\ \midrule
No Defense & 63.8$\pm$5.85 & 86.1$\pm$5.21 & 24.4$\pm$8.94 \\
Median & 56.8$\pm$7.23 & 89.6$\pm$2.49 & 21.2$\pm$8.71 \\
Trimmed Mean & 66.2$\pm$5.12 & 87.9$\pm$3.97 & 27.2$\pm$8.25 \\
Multi Krum & 73.0$\pm$3.78 & 92.6$\pm$1.42 & 35.9$\pm$3.10 \\
FoolsGold & 24.9$\pm$10.72 & 41.9$\pm$17.53 & 1.3$\pm$1.60 \\
Norm Bound & 63.5$\pm$4.45 & 86.6$\pm$7.05 & 24.1$\pm$8.86 \\
RFA & 71.5$\pm$2.66 & 92.4$\pm$1.06 & 36.3$\pm$1.12 \\
ResidualBase & 70.3$\pm$3.95 & 91.8$\pm$1.38 & 30.5$\pm$8.23 \\
\model{} & 74.4$\pm$2.85 & 93.2$\pm$0.57 & 34.9$\pm$2.18 \\ \bottomrule
\end{tabular}}
\caption{Comparison of defense performance over three datasets under label flipping attack scenarios with pollution ratio $\gamma_p=1.0$. Mean and standard deviation over ten last rounds are reported.} 
\label{Tab:untargeted_flipping_appendixv2}
\end{table*}

\begin{table*}[t!]
\centering
\scalebox{1.15}{
\begin{tabular}{l|ccc}
\toprule
Method       & CIFAR-10 & SVHN & TinyImageNet
 \\ \midrule
No Defense & 32.7$\pm$4.18 & 47.8$\pm$8.72 & 2.1$\pm$1.09 \\
Median & 67.8$\pm$4.30 & 91.5$\pm$1.21 & 28.8$\pm$3.44 \\
Trimmed Mean & 55.6$\pm$4.38 & 72.5$\pm$9.72 & 12.1$\pm$5.63 \\
Multi Krum & 52.8$\pm$5.86 & 68.4$\pm$13.72 & 15.0$\pm$4.55 \\
FoolsGold & 13.9$\pm$4.13 & 6.7$\pm$0.00 & 0.5$\pm$0.08 \\
Norm Bound & 28.2$\pm$2.49 & 42.9$\pm$10.39 & 1.2$\pm$0.67 \\
RFA & 72.0$\pm$2.85 & 92.2$\pm$0.49 & 35.8$\pm$0.80 \\
ResidualBase & 74.6$\pm$2.11 & 93.7$\pm$0.39 & 37.0$\pm$1.05 \\
\model{} & 74.8$\pm$2.42 & 93.6$\pm$0.58 & 36.1$\pm$1.37 \\\bottomrule
\end{tabular}}
\caption{Comparison of defense performance over three datasets under Gaussian noise attack scenarios. Mean and standard deviation over ten last rounds are reported.}
 \label{Tab:untargeted_gaussian_appendix}
\end{table*}

%% file: main.bbl
\begin{thebibliography}{10}\itemsep=-1pt

\bibitem{bagdasaryan2020backdoor}
Eugene Bagdasaryan, Andreas Veit, Yiqing Hua, Deborah Estrin, and Vitaly
  Shmatikov.
\newblock How to backdoor federated learning.
\newblock In {\em Proceedings of AISTATS}, pages 2938--2948. PMLR, 2020.

\bibitem{baruch2019little}
Gilad Baruch, Moran Baruch, and Yoav Goldberg.
\newblock A little is enough: Circumventing defenses for distributed learning.
\newblock In {\em Advances in NeurIPS}, volume~32, 2019.

\bibitem{blanchard2017machine}
Peva Blanchard, El~Mahdi El~Mhamdi, Rachid Guerraoui, and Julien Stainer.
\newblock Machine learning with adversaries: Byzantine tolerant gradient
  descent.
\newblock In {\em Advances in NeurIPS}, volume~30, 2017.

\bibitem{chen2017targeted}
Xinyun Chen, Chang Liu, Bo Li, Kimberly Lu, and Dawn Song.
\newblock Targeted backdoor attacks on deep learning systems using data
  poisoning.
\newblock {\em arXiv preprint arXiv:1712.05526}, 2017.

\bibitem{fang2020local}
Minghong Fang, Xiaoyu Cao, Jinyuan Jia, and Neil Gong.
\newblock Local model poisoning attacks to $\{$Byzantine-Robust$\}$ federated
  learning.
\newblock In {\em Proceedings of USENIX Security}, pages 1605--1622, 2020.

\bibitem{frankle2019lottery}
Jonathan Frankle and Michael Carbin.
\newblock The lottery ticket hypothesis: Finding sparse, trainable neural
  networks.
\newblock In {\em Proceedings of ICLR}, 2019.

\bibitem{fu2019attack}
Shuhao Fu, Chulin Xie, Bo Li, and Qifeng Chen.
\newblock Attack-resistant federated learning with residual-based reweighting.
\newblock {\em arXiv preprint arXiv:1912.11464}, 2019.

\bibitem{fung2020limitations}
Clement Fung, Chris~JM Yoon, and Ivan Beschastnikh.
\newblock The limitations of federated learning in sybil settings.
\newblock In {\em Proceedings of RAID}, 2020.

\bibitem{gu2017badnets}
Tianyu Gu, Brendan Dolan-Gavitt, and Siddharth Garg.
\newblock Badnets: Identifying vulnerabilities in the machine learning model
  supply chain.
\newblock {\em arXiv preprint arXiv:1708.06733}, 2017.

\bibitem{han2022fedx}
Sungwon Han, Sungwon Park, Fangzhao Wu, Sundong Kim, Chuhan Wu, Xing Xie, and
  Meeyoung Cha.
\newblock Fedx: Unsupervised federated learning with cross knowledge
  distillation.
\newblock In {\em Proceedings of ECCV}, pages 691--707, 2022.

\bibitem{krizhevsky2009learning}
Alex Krizhevsky, Geoffrey Hinton, et~al.
\newblock Learning multiple layers of features from tiny images.
\newblock 2009.

\bibitem{le2015tiny}
Ya Le and Xuan Yang.
\newblock Tiny imagenet visual recognition challenge.
\newblock {\em Stanford CS 231N}, 7(7):3, 2015.

\bibitem{lee2019snip}
Namhoon Lee, Thalaiyasingam Ajanthan, and Philip Torr.
\newblock Snip: Single-shot network pruning based on connection sensitivity.
\newblock In {\em Proceedings of ICLR}, 2019.

\bibitem{li2021model}
Qinbin Li, Bingsheng He, and Dawn Song.
\newblock Model-contrastive federated learning.
\newblock In {\em Proceedings of CVPR}, pages 10713--10722, 2021.

\bibitem{liu2020reflection}
Yunfei Liu, Xingjun Ma, James Bailey, and Feng Lu.
\newblock Reflection backdoor: A natural backdoor attack on deep neural
  networks.
\newblock In {\em Proceedings of ECCV}, pages 182--199. Springer, 2020.

\bibitem{lyu2020threats}
Lingjuan Lyu, Han Yu, and Qiang Yang.
\newblock Threats to federated learning: A survey.
\newblock {\em arXiv preprint arXiv:2003.02133}, 2020.

\bibitem{mcmahan2017communication}
Brendan McMahan, Eider Moore, Daniel Ramage, Seth Hampson, and Blaise~Aguera y
  Arcas.
\newblock Communication-efficient learning of deep networks from decentralized
  data.
\newblock In {\em Proceedings of AISTATS}, pages 1273--1282, 2017.

\bibitem{netzer2011reading}
Yuval Netzer, Tao Wang, Adam Coates, Alessandro Bissacco, Bo Wu, and Andrew~Y
  Ng.
\newblock Reading digits in natural images with unsupervised feature learning.
\newblock 2011.

\bibitem{park2023feddefender}
Sungwon Park, Sungwon Han, Fangzhao Wu, Sundong Kim, Bin Zhu, Xing Xie, and
  Meeyoung Cha.
\newblock Feddefender: Client-side attack-tolerant federated learning.
\newblock {\em arXiv preprint arXiv:2307.09048}, 2023.

\bibitem{pillutla2022robust}
Krishna Pillutla, Sham~M Kakade, and Zaid Harchaoui.
\newblock Robust aggregation for federated learning.
\newblock {\em IEEE Transactions on Signal Processing}, 70:1142--1154, 2022.

\bibitem{selvaraju2017grad}
Ramprasaath~R Selvaraju, Michael Cogswell, Abhishek Das, Ramakrishna Vedantam,
  Devi Parikh, and Dhruv Batra.
\newblock Grad-cam: Visual explanations from deep networks via gradient-based
  localization.
\newblock In {\em Proceedings of ICCV}, pages 618--626, 2017.

\bibitem{shafahi2018poison}
Ali Shafahi, W~Ronny Huang, Mahyar Najibi, Octavian Suciu, Christoph Studer,
  Tudor Dumitras, and Tom Goldstein.
\newblock Poison frogs! targeted clean-label poisoning attacks on neural
  networks.
\newblock In {\em Advances in NeurIPS}, volume~31, 2018.

\bibitem{steinhardt2017certified}
Jacob Steinhardt, Pang Wei~W Koh, and Percy~S Liang.
\newblock Certified defenses for data poisoning attacks.
\newblock In {\em Advances in NeurIPS}, volume~30, 2017.

\bibitem{sun2019can}
Ziteng Sun, Peter Kairouz, Ananda~Theertha Suresh, and H~Brendan McMahan.
\newblock Can you really backdoor federated learning?
\newblock {\em arXiv preprint arXiv:1911.07963}, 2019.

\bibitem{tran2018spectral}
Brandon Tran, Jerry Li, and Aleksander Madry.
\newblock Spectral signatures in backdoor attacks.
\newblock {\em Advances in NeurIPS}, 31, 2018.

\bibitem{wang2021field}
Jianyu Wang, Zachary Charles, Zheng Xu, Gauri Joshi, H~Brendan McMahan, Maruan
  Al-Shedivat, Galen Andrew, Salman Avestimehr, Katharine Daly, Deepesh Data,
  et~al.
\newblock A field guide to federated optimization.
\newblock {\em arXiv preprint arXiv:2107.06917}, 2021.

\bibitem{wu2022fedattack}
Chuhan Wu, Fangzhao Wu, Tao Qi, Yongfeng Huang, and Xing Xie.
\newblock Fedattack: Effective and covert poisoning attack on federated
  recommendation via hard sampling.
\newblock In {\em Proceedings of ACM SIGKDD}, 2022.

\bibitem{xia2021robust}
Xiaobo Xia, Tongliang Liu, Bo Han, Chen Gong, Nannan Wang, Zongyuan Ge, and Yi
  Chang.
\newblock Robust early-learning: Hindering the memorization of noisy labels.
\newblock In {\em Proceedings of ICLR}, 2021.

\bibitem{xiao2012adversarial}
Han Xiao, Huang Xiao, and Claudia Eckert.
\newblock Adversarial label flips attack on support vector machines.
\newblock In {\em Proceedings of ECAI}, pages 870--875. IOS Press, 2012.

\bibitem{xie2020dba}
Chulin Xie, Keli Huang, Pin-Yu Chen, and Bo Li.
\newblock Dba: Distributed backdoor attacks against federated learning.
\newblock In {\em Proceedings of ICLR}, 2020.

\bibitem{xie2018generalized}
Cong Xie, Oluwasanmi Koyejo, and Indranil Gupta.
\newblock {Generalized Byzantine-tolerant SGD}.
\newblock {\em arXiv preprint arXiv:1802.10116}, 2018.

\bibitem{yin2018byzantine}
Dong Yin, Yudong Chen, Ramchandran Kannan, and Peter Bartlett.
\newblock Byzantine-robust distributed learning: Towards optimal statistical
  rates.
\newblock In {\em Proceedings of ICML}, pages 5650--5659, 2018.

\end{thebibliography}
